\documentclass[11pt,a4paper]{article}
\usepackage[hyperref]{acl2020}
\usepackage{times}
\usepackage{latexsym}
\usepackage{tabularx}
\usepackage{booktabs}
\usepackage{graphicx}

\usepackage{microtype}

\aclfinalcopy 

\title{ITTC @ TREC 2021 Clinical Trials Track}

\author{Hung-Thinh Truong\textsuperscript{1} \qquad Yulia Otmakhova\textsuperscript{1} \qquad 
  Rahmad Mahendra\textsuperscript{2} \\
 \qquad \textbf{Timothy Baldwin\textsuperscript{1}}  \qquad \textbf{Jey Han Lau\textsuperscript{1}}  \qquad
  \textbf{Trevor Cohn\textsuperscript{1}} \qquad \\ \textbf{Lawrence Cavedon\textsuperscript{2}} \qquad \textbf{Damiano Spina\textsuperscript{2}} \qquad
    \textbf{Karin Verspoor\textsuperscript{$\dagger$2,1}}\\[2ex]
\textsuperscript{1}The University of Melbourne, School of Computing and Information Systems \\
  \textsuperscript{2}RMIT University, School of Computing Technologies  \\
  \textsuperscript{$\dagger$}Contact: \texttt{karin.verspoor@rmit.edu.au}   }

\begin{document}
\maketitle
\begin{abstract}
This paper describes the submissions of the Natural Language Processing (NLP) team from the Australian Research Council Industrial Transformation Training Centre (ITTC) for Cognitive Computing in Medical Technologies to the TREC 2021 Clinical Trials Track.
The task focuses on the problem of matching eligible clinical trials to topics constituting a summary of a patient's admission notes.
We explore different ways of representing trials and topics using NLP techniques, and then use a common retrieval model to generate the ranked list of relevant trials for each topic.
The results from all our submitted runs are well above the median scores for all topics, but there is still plenty of scope for improvement. 

\end{abstract}

\section{Introduction}
The TREC 2021 Clinical Trials Track formulates a patient-trial matching problem. 
The main motivation of this task lies in building an automated system to help clinical trials meet their recruitment targets, by suggesting trials relevant to a given patient.
The task consisted of retrieving relevant clinical trials from a dataset of 350K clinical trials descriptions for each of 75 queries/topics in the form of patient descriptions/clinical notes.
Each topic-trial pair is judged using three labels: \textit{eligible} (relevant and the patient does not satisfy any exclusion criteria), \textit{excluded} (relevant but the patient satisfies some of the exclusion criteria), and \textit{not relevant}.  
The analysis below reveals some challenges related to the text representation of topics and trials.

\subsection{Clinical trials}

Clinical trials are documents describing the protocols and relevant patient characteristics of a clinical research study.
Description of clinical trials can be quite long, but a core aspect of the trial description is the patient inclusion/exclusion criteria, specifying what types of characteristics or conditions a patient must have/not have in order to be suitable for the trial.

Unlike in typical retrieval tasks, for this task it is not enough to retrieve the most relevant documents based on lexical or semantic similarity. A key aspect of this task  is the patient's eligibility for a trial. Specifically, a clinical trial can be relevant for the patient based on his disease or condition, but they might be ruled out for it due to their age, gender, comorbidities (additional diseases), clinical parameters, a history of alcohol or drug use, and other such factors. Thus, apart from a high relevance of the main semantic content (namely, that a trial addresses a medical condition relevant to the patient), the retrieved clinical trial should be as close as possible to the query/topic in terms of its inclusion criteria and simultaneously as distant as possible in terms of exclusions. Another challenge is that there is also a complicated relation between inclusion/exclusion criteria and negation. For example, if a trial is looking for non-smokers, it can be expressed both as \textit{do not smoke} in the inclusion criteria and as \textit{smoke} in the exclusion criteria. Thus apart from determining the absence or presence of negation we also have to correctly match it to the exclusion or inclusion statements to accurately determine the eligibility.


\subsection{Topics}
Topics consist of synthetic patient cases created by domain experts, formulating an admission statement in an electronic health record of a patient.
Patient descriptions are generally difficult to process due to such properties of clinical text as the abundance of abbreviations and shorthand notations, incomplete sentences, medical slang, etc. They also lack structure or a particular flow, unlike biomedical texts, so it is impossible to specify particular regions of text which are more relevant than others, while some critical pieces of information (such as diagnosis) can be implied and not mentioned directly (the diagnosis is omitted in 13 out of 75 patient descriptions). They are also noisy in the sense that they contain information related not only to the patient's current condition, but also to their family history, previous diagnoses, etc. which can lead to matching non-relevant clinical trials. 

However, in addition to such challenges inherent to the clinical text, the topic set for this task was constructed in such a way that made it even more difficult. Specifically, among 75 patient descriptions, there were 15 pairs of topics where the patients had very similar conditions, symptoms, or characteristics but differed in the underlying diagnosis, clinical variables, attributes, etc. For example, patients 21 and 57 have similar symptoms and clinical signs suggestive of pancreatitis. 
However, one of them does not have obstructive pancreatitis and is a heavy drinker, so he should be matched to clinical trials regarding alcoholic pancreatitis; for the other patient the scan revealed gall bladder stones, so she should be matched to clinical trials regarding obstructive pancreatitis. 

\section{Related work}
Previous TREC iterations had several tracks with a similar retrieval setting.
The TREC Clinical Decision Support (CDS) Tracks in 2014-2016 \citep{simpson2014overview} focused on retrieving relevant abstracts of scientific publications from a PubMed snapshot. In the following years, the TREC CDS Track became the TREC Precision Medicine Track \citep{roberts2017overview} which focused on retrieving evidence-based treatment literature and clinical trials.

Most similar to the TREC 2021 Clinical Trials Track is the work by \citet{koopman2016test}. The authors introduced a dataset which has relevance judgement labels for topics from TREC CDS 2014 and a set of clinical trials from a snapshot of the ClinicalTrials.gov registry.
Furthermore, the authors introduce the use of ad-hoc queries, 
which are constructed by asking domain experts to write down what they would normally use as queries when searching for potential trials that are suitable for a patient. 
Empirical results of using different baselines show that this type of queries outperform the use of full-text, or summarized text queries.

\section{Data representation and features}
\label{sec:data_rep}
\subsection{Text representation}

We experimented with several representation settings, such as using a complete text representation -- that is, all tokens in the texts -- of both the queries and the target documents,  using complete text representation only on the target (clinical trials) side, and using a keywords/entities-based representation on both sides, in which terms were selected for inclusion in the representation after document pre-processing. For the complete text representation, we applied the default tokenization and stopword removal provided by the retrieval engine (see Section~\ref{ir-framework}). 
The methods used for keyword extraction are discussed in Section~\ref{keyword_extraction}.

\subsection{Clinical trial document structure}
Clinical trials documents consist of multiple sections, including 
a \textit{Title}; \textit{Brief Summary}, a short paragraph to summarize the key aspects of the trial; \textit{Description} describing the clinical trials in detail; \textit{Eligibility}, including inclusion and exclusion criteria; relevant concepts, MeSH terms given in \textit{Conditions}, \textit{Interventions}, \textit{Keywords}; and finally some metadata about date, times, location.
In a na\"ive approach to processing these documents, no attention is paid to this structure and all tokens are extracted for the representation. However, through our experimentation,
we found that the full text of the target documents could be too noisy and thus negatively affect the retrieval.

We therefore experimented with focusing on only the \textit{Eligibility} section of these documents, which is explicitly denoted by the \textit{eligibility} tag.
We found that we were able to retrieve more relevant documents when we used only text from this section as the target document. However, as the disease or condition name is often not mentioned in the \textit{Eligibility} section, we were getting poor results for some topics. Thus for all of our experiments 
we used titles of the clinical trials together with their eligibility criteria.

\subsection{Inclusion and exclusion criteria}

\begin{table*}[t]
    \centering
    \resizebox{16cm}{!}{
    \begin{tabularx}{\textwidth}{l| X}
        \textbf{Representation} & \textbf{Text}  \\
         \hline
        Complete text & \underline{Inclusion}: \textit{Patients $>$ or = to 18 years of age.
                                Presenting with aortic stenosis and to undergo elective aortic valve replacement or repair with or without aortic aneurysm dilation repair.
                                Able to sign informed consent document}.
                                \newline
                    \underline{Exclusion}: \textit{Patients unable to provide informed consent for any reason.
                                Patients with predominant aortic regurgitation valve disease.
                                Patients with other known connective tissue disorders (such as Marfan's Syndrome, Ehlers-Danlos Syndrome)} \\
        \hline
        Entities & \underline{Inclusion}: \textit{aortic stenosis\_\_elective aortic valve replacement\_\_neg:aortic aneurysm dilation repair}
                    \newline
                    \underline{Exclusion}: \textit{predominant aortic regurgitation valve disease\_\_connective tissue disorders\_\_Marfan's Syndrome\_\_Ehlers-Danlos Syndrome}\\
        \hline
        Concatenated & \textit{aortic stenosis\_\_elective aortic valve replacement\_\_neg:aortic aneurysm dilation repair\_\_neg:predominant aortic regurgitation valve disease\_\_neg:connective tissue disorders\_\_neg:Marfan's Syndrome\_\_neg:Ehlers-Danlos Syndrome} \\ 
    \end{tabularx}
}
    \caption{Examples of different documents representations (Inclusion and Exclusion denote separate vectors)}
    \label{tab:example}
\end{table*}

To extract inclusion and exclusion criteria, we rely on the use of regular expressions. Typically, both the inclusion and exclusion criteria are inside the \textit{Eligibility} section of a clinical trial protocol, and are explicitly denoted by a heading of either \textit{Inclusion} or \textit{Exclusion}. There exist a small number of trials where there is neither an eligibility section nor headings to separate inclusion and exclusion criteria. We decided to remove these trials from our retrieval set. 
To represent inclusion and exclusion in the trials, we experiment with using either the complete text or just the entities extracted from the criteria in different runs.
To use as input of a retrieval system, we either represent inclusion and exclusion as separate vectors or concatenate them together and 
reverse the affirmed/negated entities in the exclusion criteria.
Table~\ref{tab:example} illustrates different variances of trials representations.  

\subsection{Keyword extraction}
\label{keyword_extraction}
We experimented with several keyword extraction systems including a generation system, a named-entity recognition (NER) classifier, and a rule-based system.

\subsubsection*{Keyword generation}

In the patient-trial matching task, \citet{koopman2016test} suggest that using short keyword-based ``ad-hoc'' queries is more effective compared with using either a full patient description or brief summary thereof. 
In their work, ``ad-hoc'' queries for each patient description are defined as short phrases that a medical expert would issue to a search engine to search for relevant trials. 

Following this idea, we employ a keyphrase generation system with the aim of generating abstractive and novel phrases that can capture the key information of clinical trials.
To create data for this subtask, we use the provided set of clinical trials as the data source.
In particular, for each trial, we concatenate the brief summary and detailed description and use them as the input text.
As for the target keyphrases, we first extract the text in \textit{Condition}, \textit{Intervention}, and \textit{Keywords} fields of the clinical trials and split it by the separator (comma) to obtain separate keywords.
Following the One2Seq training paradigm \citep{yuan2020one}, we concatenate all the extracted keywords into one single string using the PREABS order \citep{meng2021empirical} where keywords are placed based on the frequencies of their occurrence in the input text.
We also limit the number of keyphrases to 8 per trial, resulting in a total of 375580 trial-keyphrases pairs.
We further split this data into train/validation/test sets with the ratio of 7:1:2 and fine-tune a T5-base model \citep{raffel2020exploring} for the generation task.

\subsubsection*{NER classifier}
Recent efforts in parsing clinical trials have introduced two large-scale public datasets with expert-annotated entities, each with a different annotation scheme: The Chia Dataset \citep{kury2020chia} containing 1000 trials, and the Facebook Research Dataset (FRD) \citep{tseo2020information} containing 3314 trials. 

\citet{tian2021transformer} suggests that using a RoBERTa model \citep{liu2019roberta} pre-trained with MIMIC-III clinical notes and eligibility criteria yielded the highest performance on both Chia and FRD dataset for clinical trials parsing, which includes the named entity recognition and relation extraction tasks.  
In this work, as the main goal is to extract as many entities as possible
, we combine the two datasets and fine-tune the released RoBERTa-MIMIC checkpoint\footnote{{https://github.com/uf-hobi-informatics-lab/ClinicalTransformerNER}} using this combined data.

\subsubsection*{Rule-based NER}

We use MetaMap\footnote{{\href{http://metamap.nlm.nih.gov}{metamap.nlm.nih.gov}}} \citep{aronson2010overview} 
to identify Unified Medical Language System (UMLS) \citep{lindberg1993unified} concepts, and then apply the UMLS Semantic Types to filter only concepts which are likely to be relevant for the clinical trials.  We retain only concepts corresponding to the following Semantic Types: \textit{aggp} (age group), \textit{cell} (cell), \textit{fndg} (finding), \textit{dsyn} (disease or syndrome), \textit{hops} (hazardous or poisonous substance), \textit{aapp} (amino acid/protein), \textit{lbtr} (lab or test result), \textit{orgf} (organism function), \textit{phsu} (pharmacologic function), \textit{qnco} (quantitative concept), \textit{sosy} (sign or symptom), \textit{topp} (therapeutic or preventive procedure).

We determined the suitability of Semantic Types for inclusion in the concept representation based on analysis of a randomly selected sample of patient descriptions and also on the list of entities commonly extracted from the clinical trials descriptions \citep{tseo2020information}. In particular, we retain only the 12 most commonly occurring Semantic Types in the descriptions, which also overlap with the most frequent entity types reported by \citet{tseo2020information}.  

\subsection{Negation detection}
Similar to other biomedical texts, negation plays an important role in comprehending the eligibility criteria.
Therefore, we aim to detect negation cues, and the entities that are in scope of those cues.
We apply a BERT-based negation detection model NegBERT \citep{khandelwal2020negbert} as our negation detection module.
The NegBERT model is pre-trained using the BioScope corpus \citep{vincze2008bioscope} where the texts are from radiology reports.
Although also being biomedical data, there are clear differences in text style and how negation is represented in clinical trials, which leads to the sub-optimal performance of NegBERT on this type of data. 
To alleviate this domain mismatch problem  
we use the NegEx model \citep{chapman2001simple} on the given clinical trials and topics data to create additional training data for NegBERT, with the assumption that a rule-based system would have high precision.
With the extracted entities of previous phases as the scope, we run NegEx to determine whether the entities are affirmed or negated.
We then randomly sample 10K sentences from this NegEx-tagged data to fine-tune the NegBERT model and apply it to detect negation in both clinical trials and patient descriptions.

\subsection{Metadata: age, gender, family history, clinical variables}

As choosing a correct clinical trial largely depends on matching its eligibility criteria in terms of age, gender, blood counts, etc., in some of our experiments we augment the data by extracting and standardizing such metadata. We use regular expressions to extract metadata of the following types:

\paragraph{Age:}
In our initial experiments we defined age brackets for patients based on MeSH classification: \textit{infant} ($<$ 2 years old), \textit{preschool child} (2-5), \textit{child} (5-12), \textit{adolescent} (12-19), \textit{adult} (19-65), \textit{young adult} (19-24), \textit{middle aged} (44-65), \textit{aged} (65-80), \textit{80 and over}. However, though this approach was suitable for the patient descriptions, where we could detect a specific mention of a patient's age and map it to a MeSH age group using a series of regular expressions, it was difficult to apply to clinical trials. Clinical trials often specify very particular age ranges for inclusion or exclusion criteria. In particular, in addition to such general age groups as \textit{adults} and \textit{children}, clinical trials can include patients \textit{over 50 years old}, \textit{under 34 years old} or \textit{between 30 and 45 years old}, which do not correspond to the MeSH age brackets directly. 

For this reason, we adopt the following approach: on the patient description side, in addition to the age group, we add the patient's age as a separate token, for example \textit{age:36}. On the clinical trials side, we use regular expressions to find  age range statements, and then add a series of individual age markers corresponding to this range, for example, \textit{age:35}, \textit{age:36}, \textit{age:37} etc.\ for the expression \textit{age 35 and over}. Whenever the age range corresponds to a MeSH group such as \textit{not younger than 19 years old}, we add an age group token (\textit{adult}).

\paragraph{Gender:}
On the clinical trials side, gender is consistently represented as \textit{male} and \textit{female}. Therefore we  normalize it to these values in patient description texts, mapping such synonyms as \textit{M}, \textit{man}, \textit{gentleman} etc.\ to the same concept. Whenever a gender marker has an age semantics (\textit{girl}, \textit{boy}), we also add a corresponding age variable.

\paragraph{Clinical variables:}
Clinical variables such as blood counts, respiratory rate, BMI, etc. pose a similar problem to the one discussed above: whereas there will be a single value for these variables in a patient description, the clinical trials usually specify a range of values in their eligibility criteria. Moreover, we noticed that though some patient descriptions differ in values of some clinical variables, 
the meaning behind these variables does not change as they still stay in the same range of normal or abnormal values. For example, though the actual values for such clinical variables as hemoglobin, white blood count and platelet count are different for topics 24 and 60 (13.5 g/dl vs 13.6 g/dl,  135000/mm3 vs 133000/mm3, and  350000/ml vs 370000/ml respectively), they all correspond to normal ranges for these variables. Thus we hypothesize that it is more important to match the semantics of the clinical variable rather than its particular value. To this end, we collect the ranges of normal and abnormal values for the major clinical variables mentioned in the patient descriptions, using medical reference sources, and then map the specific values and ranges to such tokens as \textit{creatinine:abnornal} or \textit{bp:normal}. For variables where an abnormal range has a common medical name, we also add it as a token to improve recall (for example, we use \textit{tachypnea} in addition to \textit{rr:abnormal} to represent the expression \textit{respiratory rate: 37 breaths/min}).

\paragraph{Family history:}
As our preliminary retrieval results showed that there is often confusion between the diagnosis of the patient and their family history, we find sentences mentioning family history using such expressions as \textit{family history}, \textit{father} etc and prepend the entities/keywords found in such sentences with \textit{family} (i.e. \textit{family:hypertension}).

\section{Information Retrieval framework}
\label{ir-framework}
All submitted runs were based on BM25 algorithm \citep{robertson1995okapi} using PyTerrier implementation \citep{macdonald2021pyterrier}.
We try using different combinations of topic and trial representations as discussed in Section~\ref{sec:data_rep} as input queries/documents for the retrieval framework.
We use the default configuration for BM25 with $b=0.75$, $k_1 = 0.75$ and retrieve the top 1000 documents for each query.

\section{Experiments}

\subsection{Run 1: Entities and clinical concepts}
In this first run, we only include the entities and keywords extracted as described in Section~\ref{keyword_extraction}. 
We represent both topics and trials using extracted entities.
In particular, entities are first extracted using the NER classifier and MetaMap. Then they are passed through the negation detection model to determine if they are affirmed or negated. 
In this work, we prepend negated entities with a special token ``neg:''.
It should be noted that for the exclusion criteria, we reverse the negated/affirmed status of the entities and combine them in the same list with the inclusion criteria. 
For example, if a clinical trial is excluding non-smokers, the \textit{neg:smokes} entity in the \textit{Exclusion} criteria will be changed to \textit{smoke} entity before adding it into the keywords list.
Finally, all entities are concatenated into a single string to represent a trial or topic.

\subsection{Run 2: Tokens + entities}
This run is where we use the most information on both the trials and the topics side. 
In particular, on the trials side, we use the title, the complete text of inclusion criteria, extracted entities in the inclusion and exclusion criteria, and the additional metadata such as age, gender, and clinical variable ranges.
On the patient side, the query vectors also consist of the extracted entities and metadata.
Following the strategy of the first run, we also apply negation detection, reverse the negation labels if the entities are in the exclusion criteria, and then concatenate everything into a single string to represent the trials and topics.

\subsection{Run 3: Inclusions vs.\ exclusions}
In this run, we use the complete text of patient descriptions as the query and build two indices for the trial documents representation. The first index is created using tokenized words from inclusion criteria, while the second index is based on exclusion criteria.
Suppose $S_{incl}$ and $S_{excl}$ are the sets of documents retrieved by running the query using the first and second indices, respectively. We then apply set difference operation. For this run, the intuition is to remove all the trials in the inclusion set that are also in the exclusion set. In other words, the final result in this run is the documents in the set of $S_{incl} - S_{excl}$.

\subsection{Run 4: Run 3, Entity-focused}
This run is the modification of Run 3 for query representation. Instead of using complete text, we utilize entity information to construct the query. We extract the entities from complete text of patient descriptions and then concatenate them into a single string.

\subsection{Run 5: Run 3, Entity-focused + negation}
In the last run, we modify Run 4 for query representation. Besides extracting entities from the text, we also apply negation detection. If an entity is negated, we simply exclude it from the query representation.

\section{Results and discussion}

The results of the submitted runs compared with a baseline (BL) are listed in Table \ref{tab:results}. For the baseline, we used a complete-text query against the complete-text documents without any additional 
metadata or pre-processing. We report the Average Precision at cutoff 10 (AP@10) macro-averaged across the queries (judging only the eligible trials rather than all relevant ones as correct) 
, the average distance (in precision points) from the median of AP@10 
for the runs submitted for the task by all teams (higher is better), the number of times the run achieved the highest AP@10 score 
among all submitted runs (higher is better), and the number of times the run failed to retrieve any clinical trials the patient was eligible for.

\begin{table}[t]
\centering
\resizebox{7.5cm}{!}
{
    \begin{tabular}{r | c | c c c c c}
    \toprule
    Run & BL & R1 & R2 & R3 & R4 & R5 \\
    \hline
    AP@10 & 0.216 & 0.183 & 0.257 & \textbf{0.284} & 0.245 & 0.273 \\
    $\Delta$Median & 0.054 & 0.021 & 0.096 & \textbf{0.127} & 0.084 & 0.111 \\
    Best & 1 & 0 & 0 & 2 & \textbf{3} & 1\\
    Failed & 22 & 26 & 16 & 17 & 21 & \textbf{15} \\
    
    \bottomrule
    \end{tabular}
}
\caption{Results of submitted runs.}
\label{tab:results}
\end{table}

Overall, all of the submitted runs performed better than the median, and, with the exception of Run 1, better than the baseline. The poor performance of Run 1, which used keywords rather than complete text both on the query and document sides, and a comparatively worse performance of Run 4 which differed from similar Runs 3 and 5 in that it used keywords on the query side, shows that the paucity of representation and, potentially, information loss when extracting and generating keywords negatively affects the retrieval performance. 

Another thing to note is that in terms of handling the interaction between inclusion and exclusion criteria, taking the set difference of the results retrieved using the title+inclusions text and exclusions text separately (Run 3) performs better than reversing the polarity of keywords (Run 2), which can be probably also explained by noise and information loss occurring during negation resolution and keyword extraction. However, excluding negated keywords from the query (Run 5) seems to be a feasible strategy that deserves further exploration. Finally, comparing Run 2 which had additional data in terms of clinical variables, age, and gender data against the complete-text baseline allows us to conclude that such metadata supports retrieval of more eligible clinical trials.

A detailed look at the top 10 retrieved documents for the queries where we perform better and worse did not reveal any specific patterns. 
We hypothesize that there are some topics that are ``harder'' due to the small numbers of corresponding relevant trials in the corpus, or the difference in terminological concepts usage between the queries and the trials.
For cases where we perform worse, we notice that the topics contain a large number of abbreviations (Topics 9, 10, 32, 53), or very few medical terms (Topic 44).
Furthermore, these worse cases are actually difficult topics, where there are very few eligible trials in the whole collection.

\section{Conclusion}
In this paper, we have presented our participating system in the TREC 2021 Clinical Trials Track.
We focus on exploring different ways of representing the topics and trials such as complete texts or keywords obtained using multiple information extraction strategies and using a simple BM25-based retrieval model to retrieve documents. We identify several strategies to handle exclusion criteria and negation such as set-subtraction, or removing negation.
The overall results from the shared task suggest that our system can be improved upon by enriching our approach with more sophisticated yet common information retrieval techniques; such as through the use of neural rerankers, and other ranking fusion strategies to handle inclusion and exclusion criteria.

\section*{Acknowledgements}
We would like to thank other members of the Natural Language Processing (NLP) group from the Australian Research Council Industrial Transformation Training Centre (ITTC) for Cognitive Computing in Medical Technologies, including Yiyuan (Gracie) Pu, Daniel Beck, Simon Šuster, and Antonio Jimeno Yepes for their contributions to discussions about the task. We acknowledge the funding support of the Australian Research Council grant  that supports the work of the ITTC.
\bibliography{anthology,acl2020}
\bibliographystyle{acl_natbib}

\end{document}